\documentclass{article}

\usepackage{arxiv}

\usepackage[utf8]{inputenc} 
\usepackage[T1]{fontenc}    
\usepackage{hyperref}       
\usepackage{url}            
\usepackage{booktabs}       
\usepackage{amsfonts}       
\usepackage{nicefrac}       
\usepackage{microtype}      
\usepackage{lipsum}		
\usepackage{graphicx}

\usepackage{amsmath}
\usepackage{algorithm}
\usepackage{algorithmicx}
\usepackage{algpseudocode}
\usepackage{subfigure}

\title{Data Augmentation for Deep Candlestick Learner}

\author{{Chia-Ying Tsao}\\
	Department of Economics\\
	National Taiwan University\\
	Taipei 10617, Taiwan\\
	\And
	{Jun-Hao Chen} \\
	Department of Computer Science \& Information Engineering\\
	National Taiwan University\\
	Taipei 10617, Taiwan\\
	\And
	Samuel Yen-Chi Chen\\
	Computational Science Initiative\\
	Brookhaven National Laboratory\\
	Upton, NY 11973, USA\\
	\And
	Yun-Cheng Tsai\\
	School of Big Data Management\\
	Soochow University\\
	Taipei 11102, Taiwan\\
	\texttt{pecutsai@gm.scu.edu.tw} \\
}

\begin{document}
\maketitle

\begin{abstract}
To successfully build a deep learning model, it will need a large amount of labeled data. However, labeled data are hard to collect in many use cases. To tackle this problem, a bunch of data augmentation methods have been introduced recently and have demonstrated successful results in computer vision, natural language and so on. For financial trading data, to our best knowledge, successful data augmentation framework has rarely been studied. Here we propose a Modified Local Search Attack Sampling method to augment the candlestick data, which is a very important tool for professional trader. Our results show that the proposed method can generate high-quality data which are hard to distinguish by human and will open a new way for finance community to employ existing machine learning techniques even if the dataset is small.
\end{abstract}


\section{Introduction}
\label{sec:introduction}
How to augment limited stock price data is an open problem in stock trend prediction. Most innovative data augmentation schemes adopted in image processing community cannot be used directly in time-series data. Although the traditional financial data simulation method can generate time-series data, there are some defects when considering the real-world market. For example, Monte Carlo simulation is one of the primary traditional tools applied extensively in financial engineering research, economics, and a wide array of other fields during the past four decades~\cite{mcleish2011monte}. However, Monte Carlo simulation is ultimately a statistical model, which means it requires several assumptions. Those assumptions may be unrealistic and depends on the individual circumstances. There are three primary disadvantages as follows:
\begin{enumerate}
    \item Monte Carlo simulations need distribution assumptions to built around a specific type of statistical distribution. If we use the right distribution assumption, the results are valid. However, if we use the wrong one then the results will be meaningless~\cite{borak2013statistics}.
    \item Monte Carlo simulations need input assumptions and are only as good as the inputs they start within. For instance, a simulation might be used to evaluate the value of a market price process. However, if the market price process unfits the real market requirements, we cannot make generalizations around the actual market situations at various points in the future~\cite{visser2007simulation}.
    \item Monte Carlo simulations need certain assumptions of the mathematical formulation of the dynamics that drive the end values. Sometimes those formulas are straightforward and undeniable. However, they are not suitable in many real market cases~\cite{hoesli2006monte}.
\end{enumerate}

Data augmentation methods based on deep learning develop rapidly in recent years. With the property of universal approximation, deep learning could directly learn the representation of data without providing specific assumptions on statistical distribution and the analytical formulation of the data~\cite{bengio2013deep}. Further, harvesting the large expressive power of deep neural networks to capture the unknown underlying dynamics of the financial data can help the model generalization. In some domains, like image classification and speech recognition, data augmentation based on deep learning has successfully provided contribution by improving accuracy~\cite{perez2017effectiveness, ko2015audio}.

Variational Autoencoder (VAE) is a commonly used data augmentation method. It performs well in many fields like image classification and speech recognition~\cite{garay2018data, nishizaki2017data, hsu2017unsupervised}. Furthermore, models like Conditional Variational Autoencoder (CVAE) could enable us to have better control on generating labeled data~\cite{NIPS2015_5775}. However, the calculation of loss function and the Gaussian encoder/decoder assumptions often reduce the effectiveness of VAE-based models in generating realistic samples~\cite{DBLP:journals/corr/ZhaoSE17a, NIPS2016_6158, dai2018diagnosing}.

Regarding the fact that the difference among financial candlestick data is subtle and the aforementioned disadvantage might mislead the classification model, we propose a novel Modified Local Search Attack Sampling model based on the previous work on explainable candlestick learner~\cite{chen2020explainable}. The model consists of two steps:
\begin{enumerate}
\item Train the GAF-CNN model~\cite{tsai2019encoding}.
\item Generate data by attacking the trained model.
\end{enumerate}
The main concept in the proposed method is that we perturbed the input so slightly that the trained GAF-CNN model will not misclassify.

To evaluate the performance of our model, we trained the Conditional Variational Autoencoder (CVAE) and Modified Local Search Attack model respectively to generate the labeled candlestick data. Next, we designed an online questionnaire and performed statistical analysis to verify our conjecture. We expect the result of our model based on Modified Local Search Attack will be more stable and perform better since the method simply adds slight perturbations to the real input data, which should be potentially more realistic compared to the ones generated from VAE-based models.

\section{Background}

When to entry and exit market is one of the most important issue in finance. Helpful information for decision making often hide in fluctuated market prices. It's hard to capture these information with bare human eyes. In 18th century, Munehisa Homma comes up with an visualize idea, candlestick chart, to conveniently observe the price of rice market~\cite{nison2001japanese}. Each bar in candlestick concentrate a series of time series pricing data to four values, open, high, low, and close price (OHLC):
\begin{enumerate}
    \item Open: the first price of the time period.
    \item High: the highest price of the time period.
    \item Low: the lowest price of the time period.
    \item Close: the last price of the time period.
\end{enumerate}

\begin{figure}[h]
\centering
\includegraphics[scale=0.45]{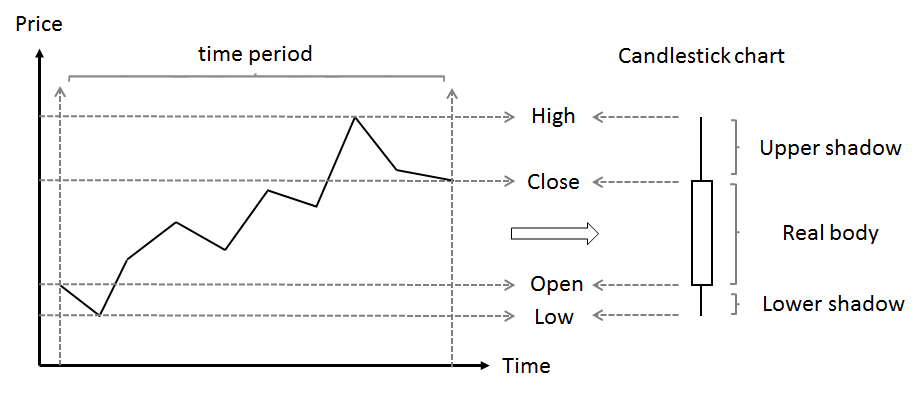}
\caption{Candlesticks display all the market needed information, such as open, high, low, and close prices.}
\label{candlestick_intro}
\end{figure}

We generally use real body and shadows to depict a bar. Real body is the area between open and close price. When open price is higher than close price, real body would be filled by black or green. In contrast, when close price is higher than open price, real body would be filled by white or red. Shadows appear the lowest and highest price during the time period. Figure~\ref{candlestick_intro} illustrates the entire structure of the candlestick chart.

\begin{figure}[ht]
\centering
\includegraphics[width=0.45\textwidth]{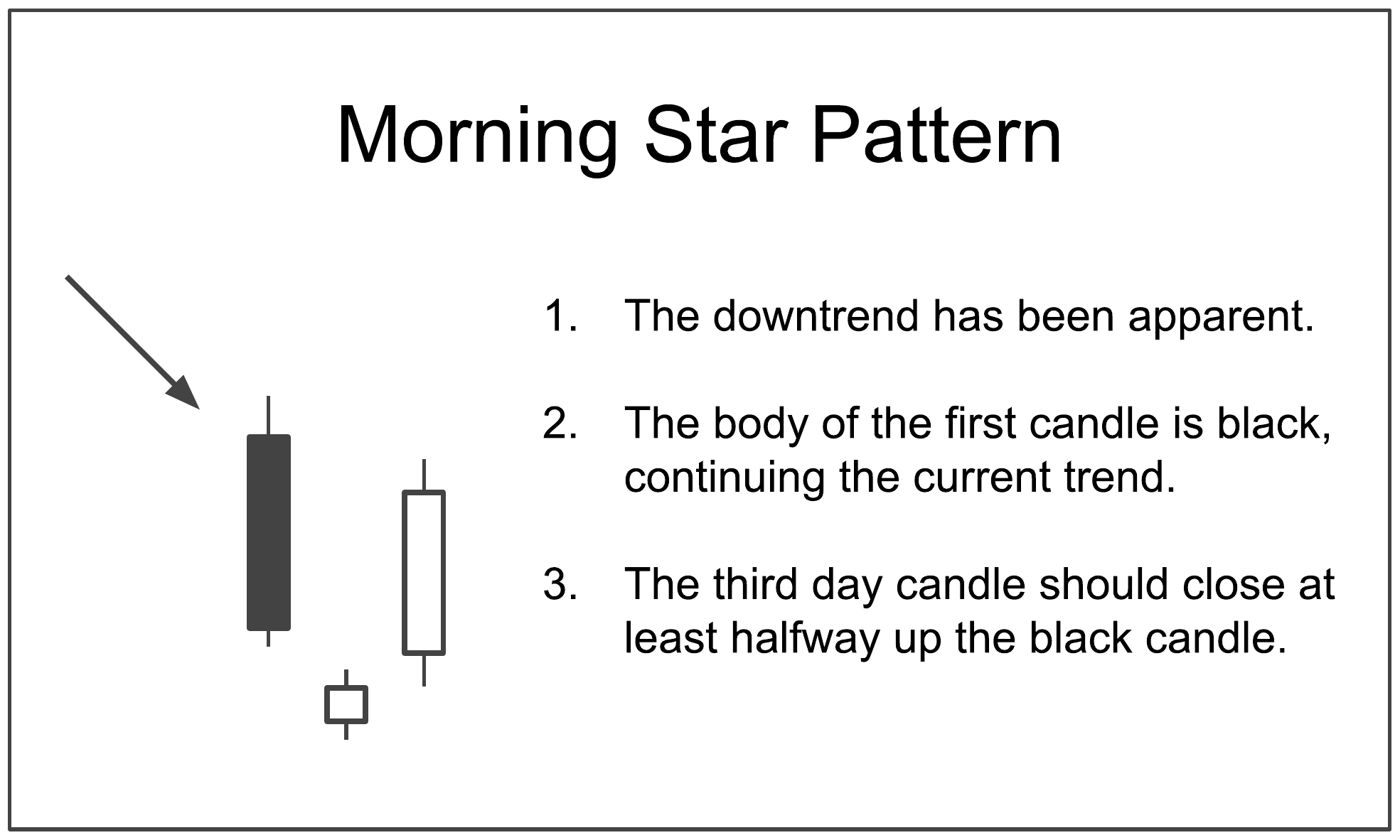}
\caption{{\bfseries Illustration of Morning Star Pattern.} The left-hand side shows the appearance of the Morning Star pattern. The right-hand side shows the critical rules of the Morning Star pattern.}
\label{morning_intro}
\end{figure}

\begin{figure}[ht]
\centering
\includegraphics[width=0.45\textwidth]{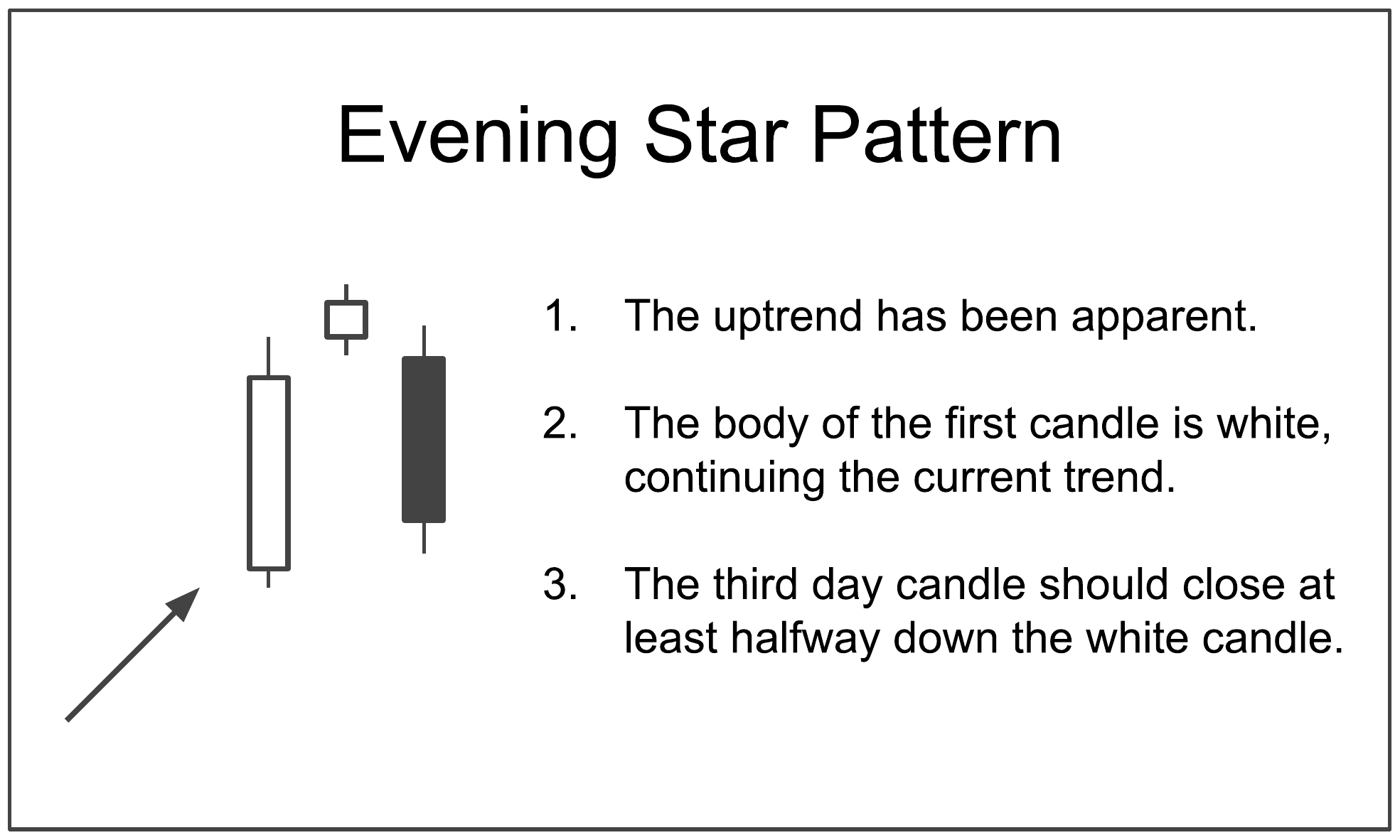}
\caption{{\bfseries Illustration of Evening Star Pattern.} The left-hand side shows the appearance of the Evening Star pattern. The right-hand side shows the critical rules of the Evening Star pattern.}
\label{evening_intro}
\end{figure}

The Major Candlesticks Signals refers to several basic patterns of candlestick, eight of these patterns are adopted in this study, including: 
\begin{enumerate}
    \item Morning Star could be recognized by a downtrend followed by a long black bar, a shorter black or white bar with a short body and long shadow, and a long white bar. The middle bar of the morning star captures a moment of market indecision where the bears begin to give way to bulls. The third bar confirms the reversal and can mark a new uptrend. Figure~\ref{morning_intro} shows the morning star based on the description.
    \item Evening Star is a bearish candlestick pattern could be recognized by uptrend followed by three bars: a long white bar, a short bodied bar, and a black bar. The pattern will be more visible with a long black bar than with a short black bar. Figure~\ref{evening_intro} shows the evening star based on the description.
    \item Bullish Engulfing occurs when a small black bar fully engulfed by a following large white bar. It appears in a downtrend.
    \item Bearish Engulfing contains a white bar followed by a large black bar that eclipses or engulfs the smaller up bar. It signals lower prices to come.
    \item Shooting Star is a bearish bar with a small real body, a long upper shadow and little or no lower shadow. It appears after an uptrend.
    \item Inverted Hammer forms a upside down hammer-shaped pattern. It act as a warning of a potential reversal upward and called a shooting star when it appears in an uptrend.
    \item Bullish Harami is a a long bar followed by a smaller bar which is completely contained within the vertical range of the previous body. It indicates the end of a bearish trend. 
    \item Bearish Harami is composed of a long white bar and a small black bar. The opening and closing prices of the second candle must be contained within the body of the first candle.
\end{enumerate}

\section{Literature review}
Data augmentation is a technique that expands the dataset based on the existing data. Certain machine learning tasks like image classification performs better when having large amount of data since models trained with small dataset often suffer from overfitting. In fields such as computer vision or natural language processing(NLP), data augmentation becomes useful and could be applied to specific problems.

There are various ways to implement data augmentation. Take image processing for example, the main techniques fall under the category of data warping, including geometric transformations, color space, rotation and so on. It creates a wider variety of samples by randomly change the parameter of the image structure. The AlexNet CNN architecture~\cite{NIPS2012_4824} which was developed by Krizhevsky et al has revolutionized image classification, and they use data augmentation to increase the size of dataset.

As for deep learning models, in addition to GANs~\cite{NIPS2014_5423}, variational autoencoders (VAEs)~\cite{kingma2013auto} occupies a dominant status. It is a likelihood-based
model that learns the latent representations. VAEs attain state-of-the-art results in semi-supervised learning~\cite{maaloe2016auxiliary}, and have been applied to draw image~\cite{gregor2015draw} or text classification~\cite{xu2017variational}. 
After the original VAE model was introduced, different kinds of its extensions show up. One of which is conditional variational autoencoders (CVAE)~\cite{NIPS2015_5775}, which could deal with conditional tasks.

\section{Methods}
\subsection{GAF-CNN}
The GAF-CNN is a two-step model that could effectively handle classification task in time series. Firstly, encode time series data into two-dimensional matrix with Gramian Angular Field (GAF) method. Secondly, use these two-dimensional matrix to train the Convolutional Neural Network (CNN) model with architectures based on the complexity of empirical task~\cite{wang2015imaging}.

The GAF is a novel time series encoding method proposed by Wang and Oates~\cite{wang2015imaging}, which represents time series data in polar coordinate system and converts angles into symmetric two-dimensional matrix. GAF contains two different forms, summation and difference version, and the summation version is adopted in this study. Each element of the GAF matrix is the cosine of the summation of angles.

The first step to making a GAF matrix is to normalize the given time series data $X$ into values between $[0, 1]$. The following equation shows the entire normalization process, where notation $\widetilde{x}_{i}$ represents the normalized data.
\begin{align}
\widetilde{x}_{i}&=\frac{x_i-\min(X)}{\max(X)-\min(X)}
\end{align}
After normalization, the second step is to represent the normalized time series data in the polar coordinate system, whose process is revealed in the following two equations.
\begin{align}
\phi&=\arccos(\widetilde{x}_i), -1\leq \widetilde{x}_i \leq 1, \widetilde{x}_i \in \widetilde{X}\\
r&=\frac{t_i}{N}, t_i\in\mathbb{N}
\end{align}
Finally, sum the angles with cosine function to make the GAF by the following equation:
\begin{align}
\textup{GAF}=\cos(\phi_i + \phi_j) = \widetilde{X}^T \cdot \widetilde{X} - \sqrt{I-\widetilde{X}^2}^T\cdot \sqrt{I-\widetilde{X}^2}
\end{align}

The GAF has two essential properties. First, the mapping from the normalized time series data to GAF is bijective when $\phi \in [0,\pi]$. The GAF matrix can be inversely transformed to time series data through its diagonal elements. Second, in contrast to Cartesian coordinates, the polar coordinates preserve absolute temporal relations. 

After GAF based encoding, data had been transformed to two-dimensional matrix which match the input format of CNN model. CNN models have been very successful in classification task, which extract various features of each class through the deep layers. With GAF encoding, the powerful capability offered by CNN could be utilized in processing time series data.

\subsection{Modified Local Search Attack Sampling}
Generally, perturbation is used to attack the model for robustness test. When it comes to the other aspect, data with small perturbation could be seen as a different data. In other words, perturbation can be applied on data augmentation. Moreover, since the perturbation boundary is controllable in Modified Local Search Attack, extreme values of generated data are controllable~\cite{chen2020explainable}.

Firstly, we perturb the diagonal elements in the GAF matrix. we then calculate the corresponding values of non-diagonal elements and output the perturbed GAF matrix.
Secondly, send this perturbed GAF into the CNN model to get the classification results. If the perturbed input is not misclassified, then we'll collect it as the generated data. Or, simply repeat the procedure described above and reset to origin every 3 episodes. The detail of the algorithm is in Algorithm~\ref{alg}.

\subsection{Conditional Variational Autoencoder}
Conditional variational autoencoder (CVAE) is an extension of VAE, which takes both data $X$ and the desired category label $Y$ as input, and generates data that conditioned on the label~\cite{NIPS2015_5775}. We use the model since it has been shown an increased diversity in the generated samples, while at the same time more stable to train.

CVAE consists of two parts. The encoder encodes the input into the latent space and learns the joint distribution based on the condition label $y$. The decoder then reproduces the input distribution from the latent variable $z$ which concatenates with $y$ to generate the optimal output. The combined model is trained by minimizing the sum of the reconstruction loss and the KL divergence term.

The architecture of CVAE in this study starts from a simple network with three dense layers in both encoder and decoder, the activation function is leaky ReLU with $\alpha = 0.2$. The training data $X$ is the candlestick data processed by GAF, reshaped to shape $(4096, )$. The label $Y$ represents the class of data, ranging from $1$ to $8$ and been shaped into $(1, )$.

\begin{figure}[h]
\begin{center}
\includegraphics[width=1.0\linewidth ]{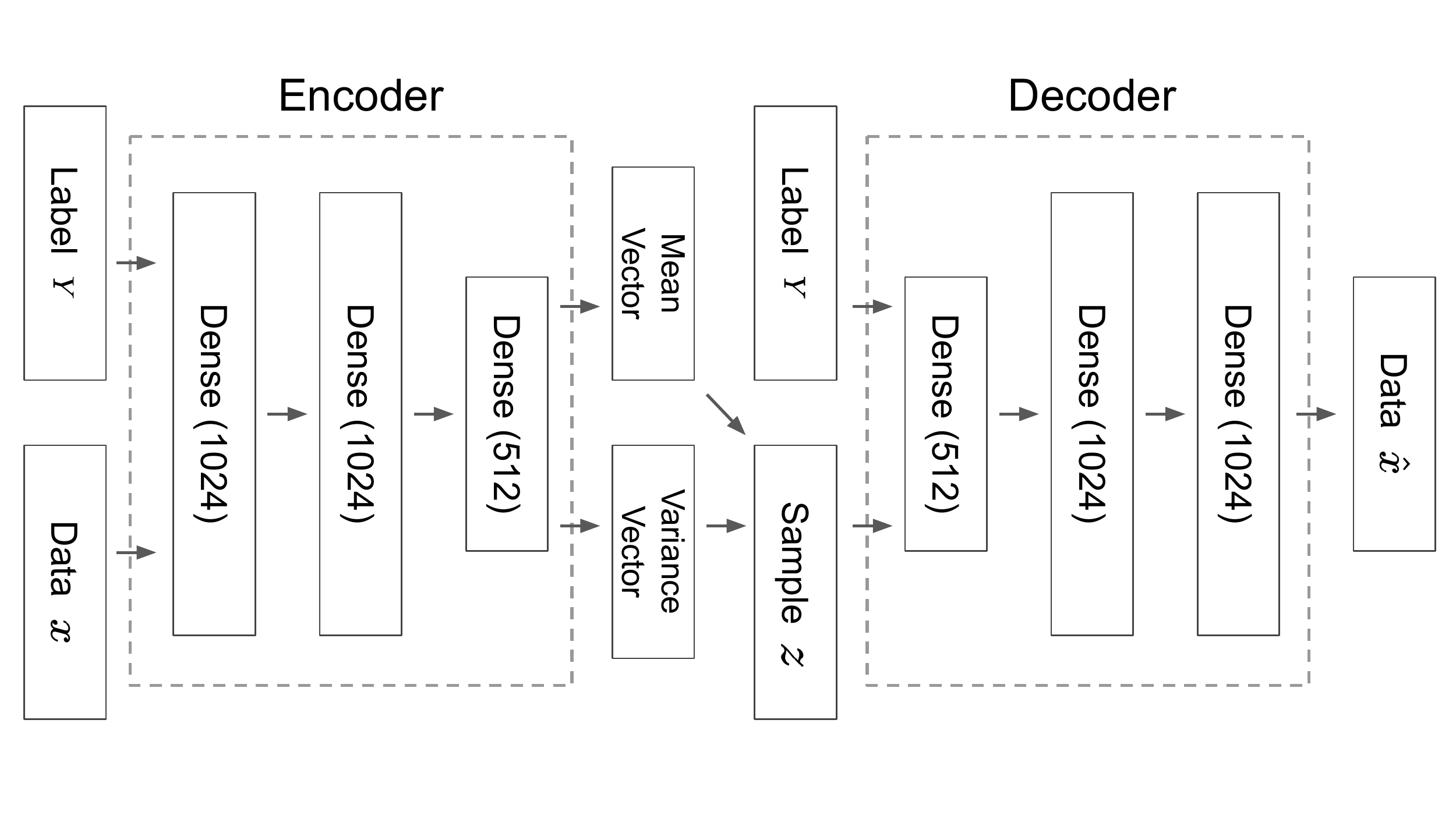}
\end{center}
\caption{The architecture of Conditional Variational Autoencoder (CVAE) in this study.}
\label{fig:cvae_model}
\end{figure}

In this work, CVAE could be used to capture the pattern feature between different candlestick labels. We use the class label of candlestick pattern as the condition label $Y$. The encoder takes price data with pattern label $Y$ as input and the decoder takes latent vector $Z$ with the label to reconstruct data.

\begin{algorithm}[tb]
\begin{algorithmic}

\State Load a single GAF two-dimensional array $A$
\State Set $T = \text{length of the time-series}$ 
\State Keep a copy of $A$ in memory $D$
\State Initialize the counter $t = 0$
\For{episode $=1,2,\ldots,R$}

\If{$t = 3$}
\State Reinitialize the $A$ to the original value from memory $D$
\State Reset the counter $t = 0$
\EndIf
\For {$l = 1,2,\ldots,T$}
	\State Sampling a random perturbation scale $r_l$ from uniform distribution $[0.99,1.01]$
	\State Calculate the perturbed result $=r_l \times A[l,l]$
	\State Set $A[l,l] = r_l \times A[l,l]$ 
\EndFor

\State $t = t + 1$
\State Recalculate the time series from perturbed $A$ and then encode into a new GAF matrix $A'$
\If{$A'$ \text{is not adversarial}}
    \State Collect $A'$ as generated data
\EndIf
\EndFor
\end{algorithmic}
\caption{Modified Local Search Attack Sampling}
\label{alg}
\end{algorithm}

\section{Experiments}

\begin{figure}[h]
\begin{center}
\includegraphics[width=0.7\linewidth ]{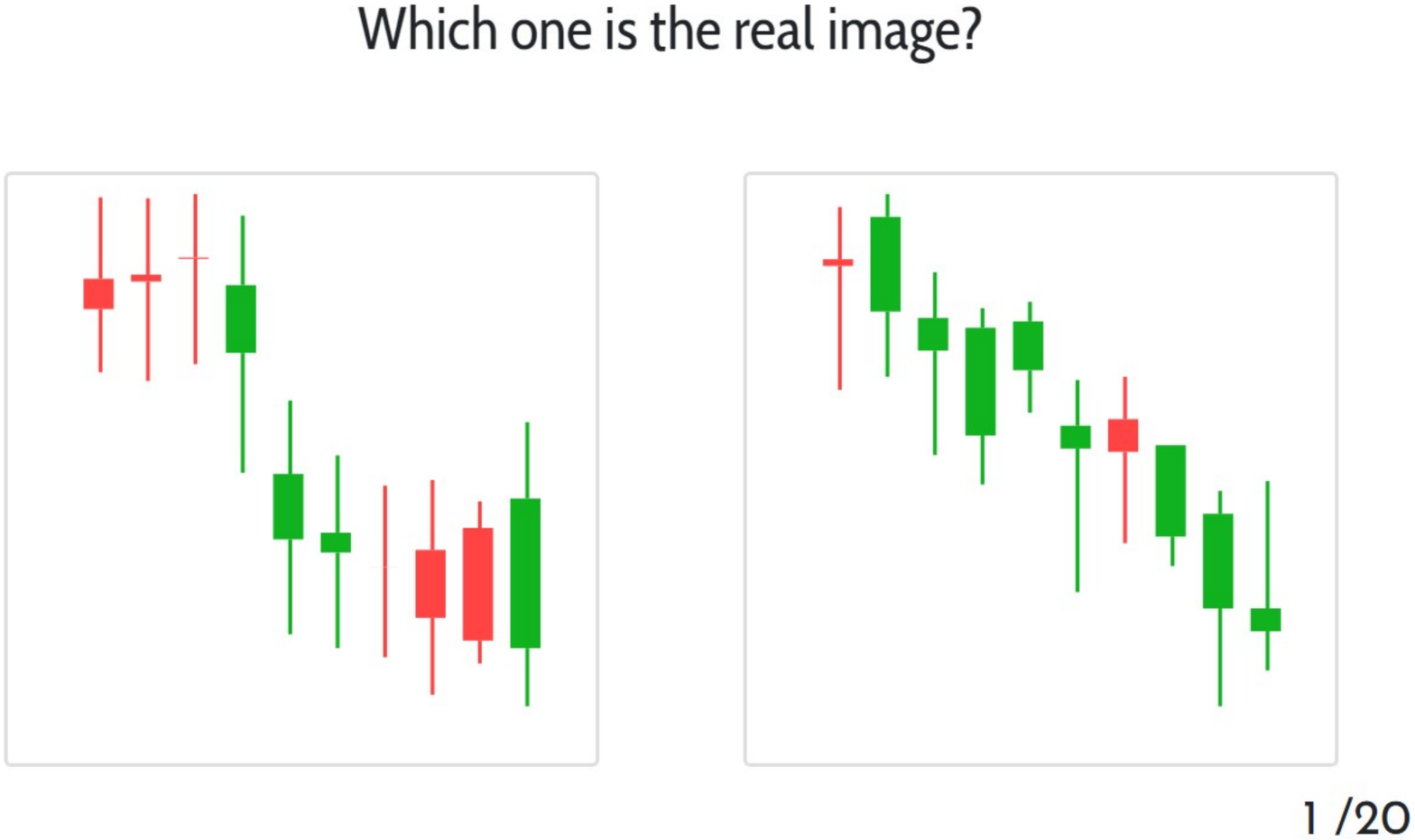}
\end{center}
\caption{The interface of the online questionnaire.}
\label{fig:website}
\end{figure}

\begin{figure}[h]
\begin{center}
\includegraphics[width=0.7\linewidth ]{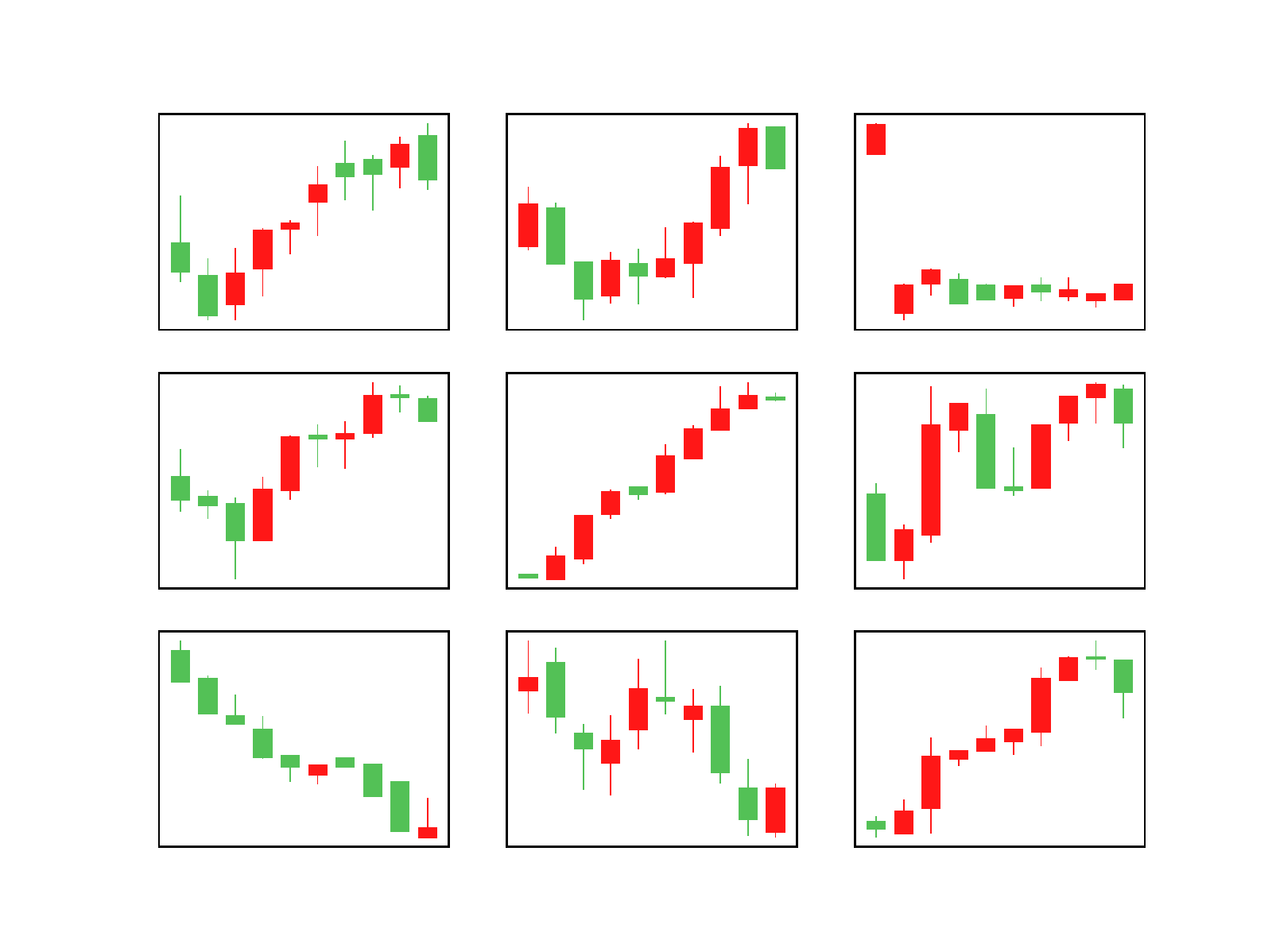}
\end{center}
\caption{The example data generated by the Modified Local Search Attack Sampling model.}
\label{fig:adv_examples}
\end{figure}

\begin{figure}[h]
\begin{center}
\includegraphics[width=0.7\linewidth]{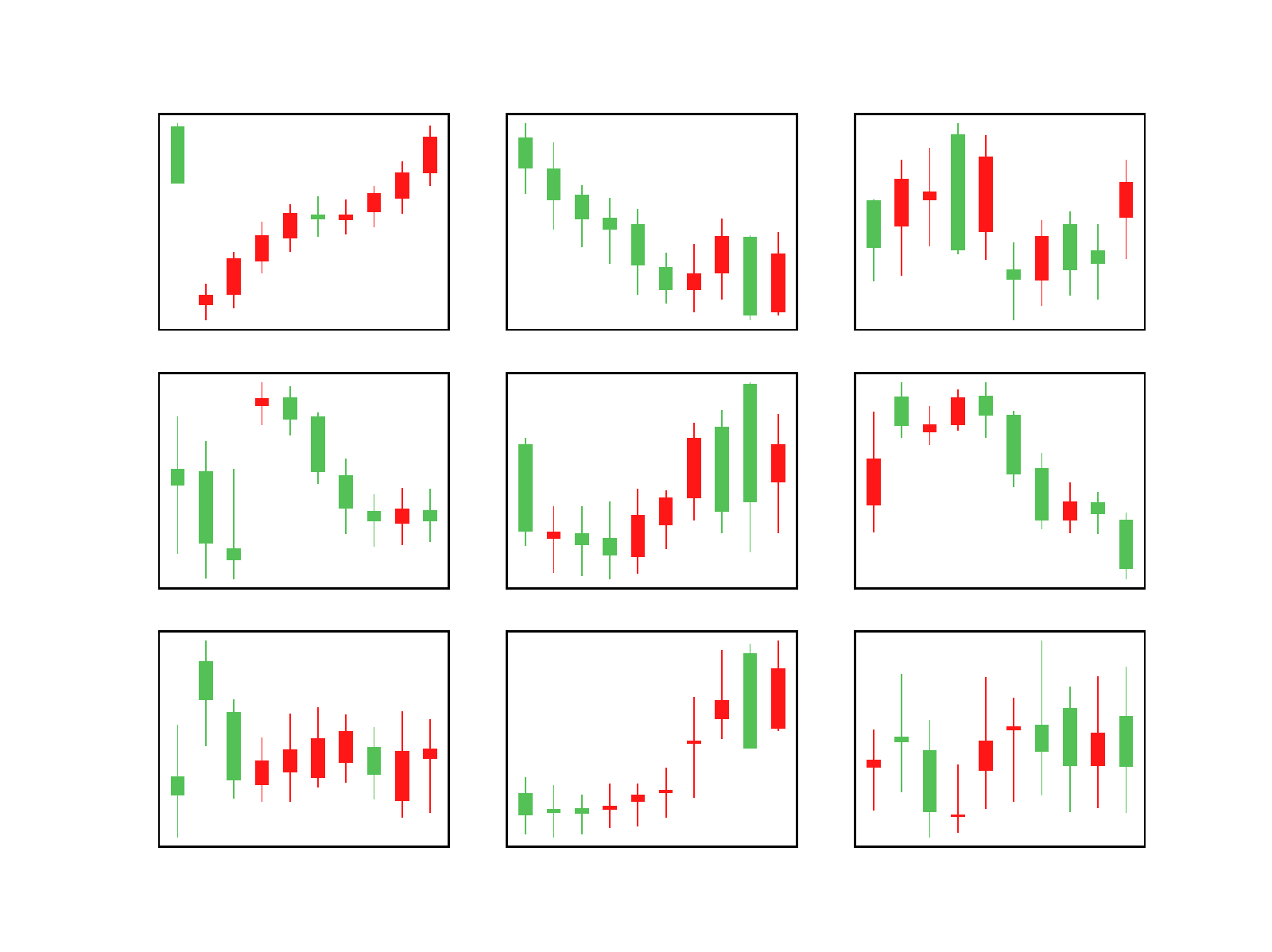}
\end{center}
\caption{The example data generated by the CVAE model.}
\label{fig:cvae_examples}
\end{figure}

\begin{figure}[h]
\begin{center}
\includegraphics[width=0.7\linewidth]{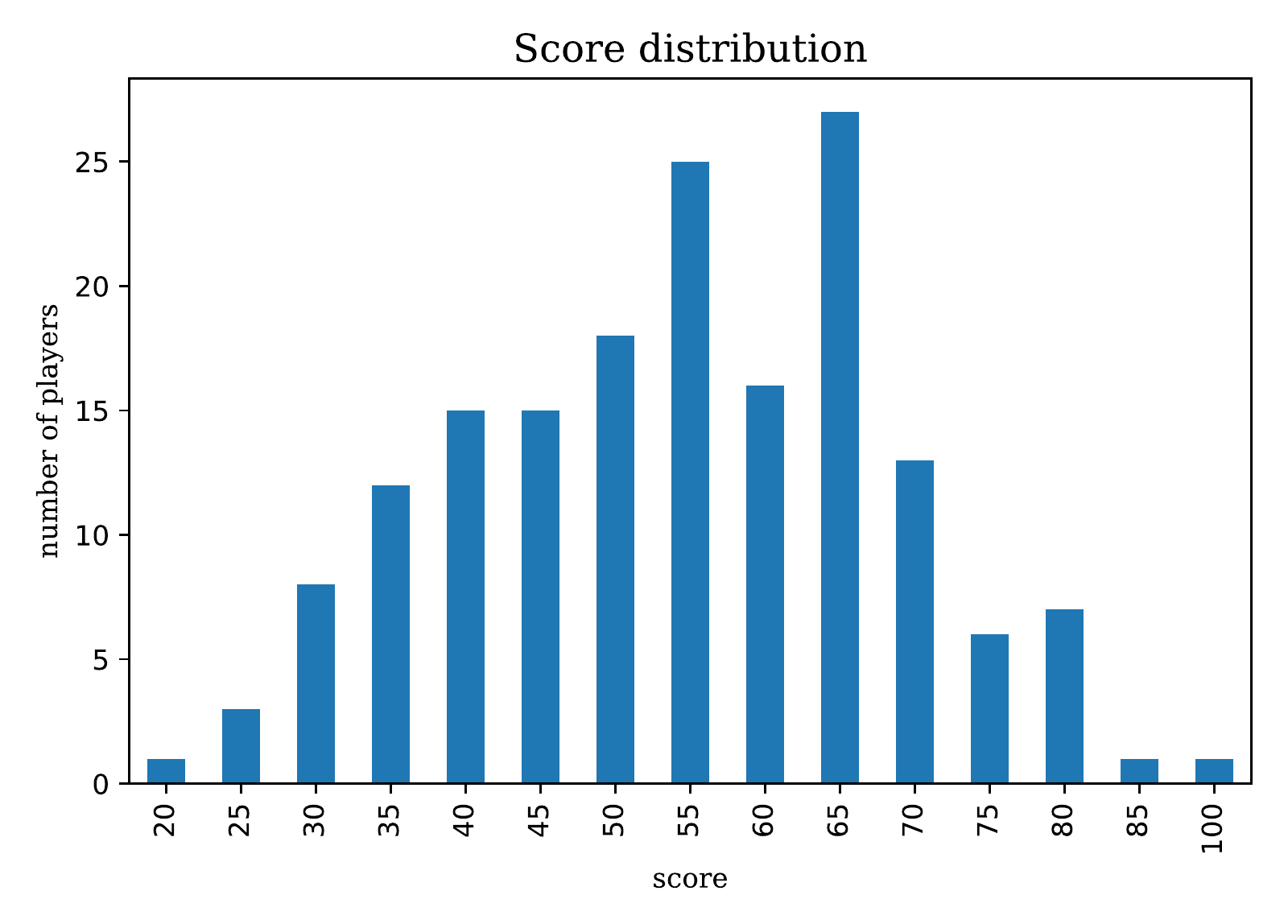}
\end{center}
\caption{The score distribution of questionnaire result.}
\label{fig:score}
\end{figure}

\begin{figure}[h]
\begin{center}
\subfigure[The first attack result example.]{\includegraphics[width=0.7\linewidth ]{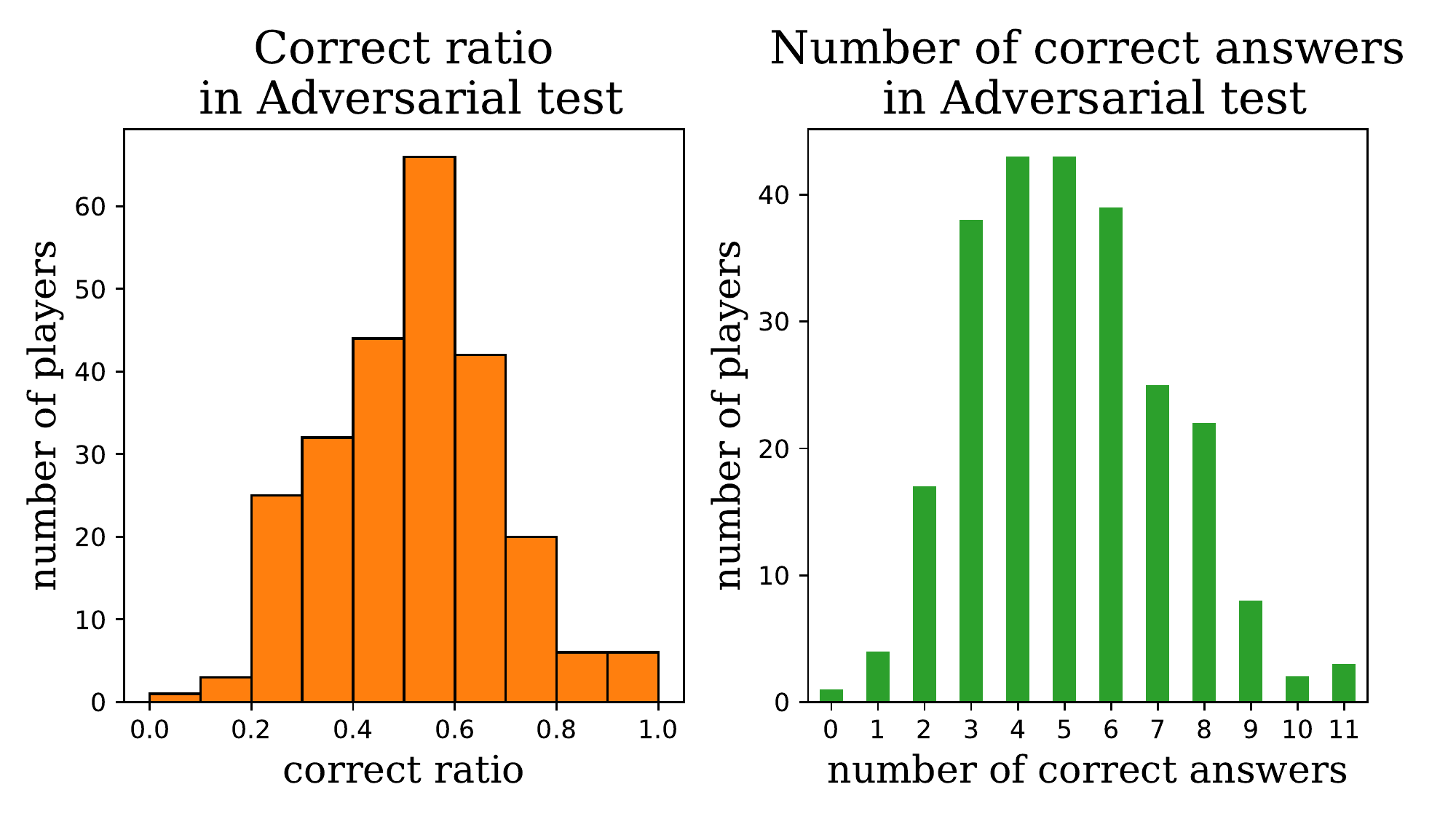}}
\subfigure[The second attack result example.]{\includegraphics[width=0.7\linewidth ]{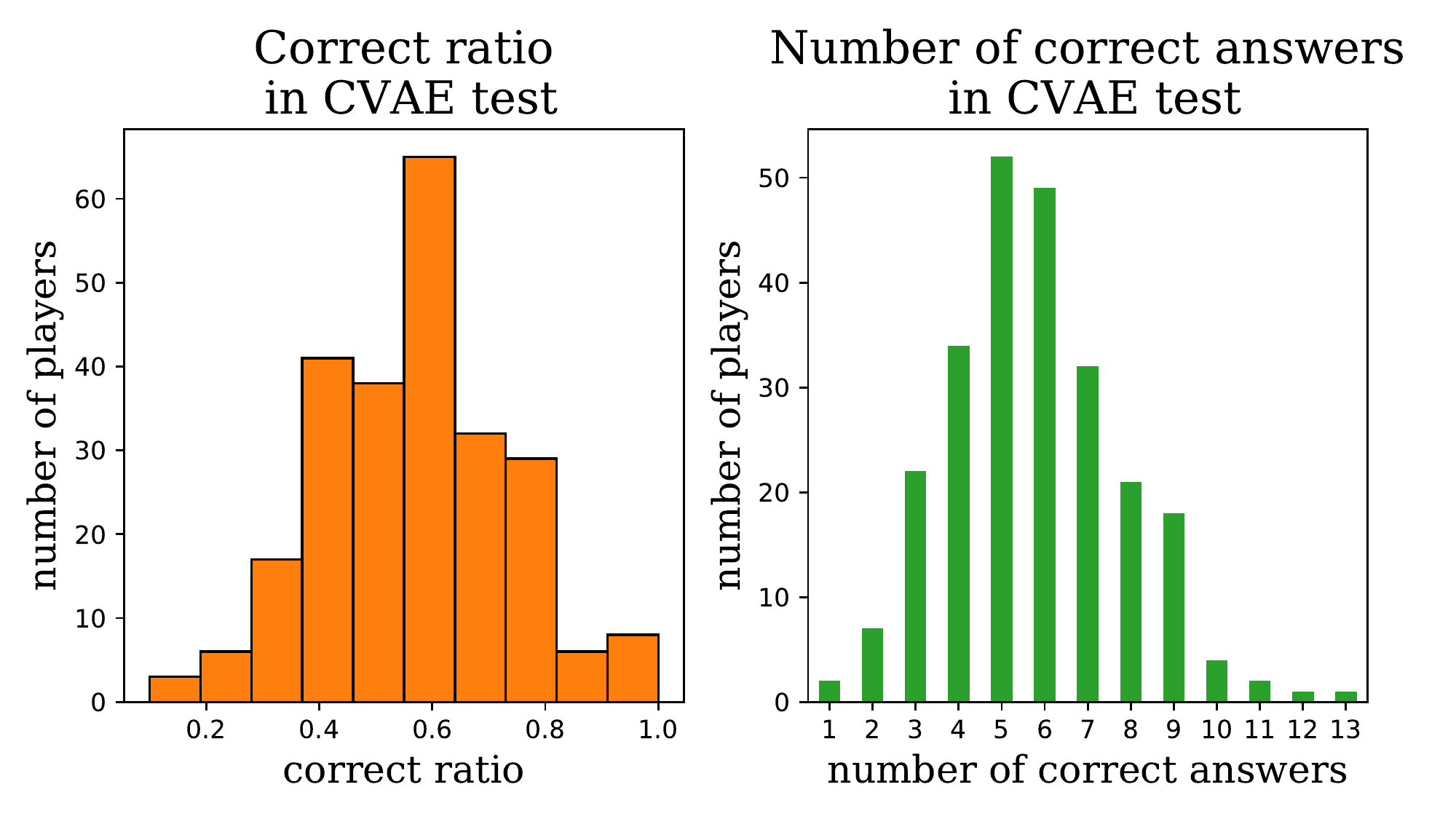}}
\end{center}
\caption{The attack result example of evening star pattern.}
\label{fig:adv_and_cvae}
\end{figure}

We use EUR/USD 1-minute open, high, low, and close price data. The training set is from January 1, 2010 to January 1, 2018, there are eight patterns and each label includes $1500$ data. To verify that the proposed Modified Local Search Attack Sampling model can produce more realistic data than average level as we expect, we apply two models to the same dataset and compare the output in each pattern. We use GAF method to encode data for both models' input.


We applied both models on the same dataset, generating $4000$ samples for each model. Meanwhile, we randomly select $4000$ samples from the real data. Each of the $4000$ samples contains an array shaped $(10, 4)$, which represents $10$ days of candlestick data in four prices (Open, High, Low, Close). Figure~\ref{fig:adv_examples} and~\ref{fig:cvae_examples} respectively show nine examples generated by Modified Local Search Attack Sampling and CVAE model.

To compare the performance of the two models, we design a web-based game as the questionnaire. The URL is\\
\url{https://ntuelvish.github.io/candlestick_challenge/v3/index.html?fbclid=IwAR34FBqGM_x-nBKonjcxMvypqzb192hJ8UbLz16pU5zBelFcvw7HKpLnIrI}.

When the player enters the game, he/she will see two images. One of which is the real data, and the other is the generative model's output data that could be from either CVAE or Modified Local Search Attack Sampling model. The player is asked to choose the one which he/she considers from real data, there are $20$ questions in total.

We record the players' information such as score, answer in each question and the testing time. Based on the situation that not all of the players pay full attention in the test, we filter the outliers to ensure the result be reliable. If the players take less than $5$ seconds playing the game or even didn't complete the test, we will drop the data. There are $245$ clean data of each model.

Finally we plot the histogram to show the distribution of the players' answers to CVAE and the Modified Local Search Attack Sampling model questions. We further perform the hypothesis testing on the collected data. Since the two models take the same input data, we suppose the two populations of the output are dependent, the dependent paired $t$-test is adopted to analyze the result. We then make the statistical inference according to this result.

\section{Results}

\begin{table}[]
\centering
\begin{tabular}{ccc} \toprule
Model       & Number of  Samples & Correct Ratio (\%) \\ \midrule
CVAE        & 2419               & 56.63              \\
Adversarial & 2364               & 51.99              \\ \bottomrule
\end{tabular}
\caption{The experimental questionnaire result between CVAE and Modified Local Search Attack Sampling model except outliers. The number of samples is values answered with valid respondents (see the main text for details). The correct ratio represents the percentage of  participants correctly recognizing the real data.}
\label{tab:questionaire_result}
\end{table}

\begin{table}[]
\small
\centering
\def\arraystretch{1.3}
\begin{tabular}{ccclll} \toprule
\multicolumn{6}{c}{Dependent Paired T Test} \\
$H_0$ & N & Mean & Std & T-value & P-value \\ \midrule
$\mu_{\textup{cvae}}=\mu_{\textup{adv}}$ & 245 & -0.0575 & 0.2386 & -3.7736 & 0.0002 \\ \bottomrule
\end{tabular}
\caption{Two sided test at the $0.05$ significance level.
The null hypothesis stated here is the average correct ratio per capita does not has significant difference between CVAE and Modified Local Search Attack Sampling data.}
\label{tab:ttest_table}
\end{table}

\subsection{descriptive statistics}
According to the questionnaire result, Figure~\ref{fig:score} shows the score distribution of questionnaire result. The distribution of score is very close to normal distribution. When the mean of scores is close to $50\%$, it means that it is very difficult to distinguish between the realistic and the generated data. The mean of scores is $54.32\%$, which means that it has a hard time to identify the realistic data.

Our result verifies that both of our models have the capability to augment financial candlestick data. To further investigate which model is better, we evaluate the correct ratio per capita to each model, the higher correct ratio means the generated data is easier to be identified, otherwise, the generated data is more difficult to be identified. Table~\ref{tab:questionaire_result} shows the correct ratio per capita of both models. The correct ratio per capita of the Modified Local Search Attack Sampling model is lower than the CVAE model. 

\subsection{hypothesis testing}
In order to reasonably inference if these models have significant difference, we perform a dependent paired $t$-test on the correct ratio per capita. In null hypothesis $H_0$, we assume that two models has no significant difference.

According to the result in Table~\ref{tab:ttest_table}, the two-tailed dependent paired $t$-statistic is $-3.7736$ and the $p$-value is $0.0002$. Since the $p$-value is less than $0.025$, this means that the average correct ratio per capita of CVAE model and Modified Local Search Attack Sampling model are significantly different. On the basis of the fact that Modified Local Search Attack Sampling model have lower mean of correct ratio, we can state that our Modified Local Search Attack Sampling model is significantly better than the CVAE model.

In short, humans tend to distinguish CVAE data from real data easier, while have difficulty recognizing data generated by Modified Local Search Attack Sampling model. It shows that the data generated based on adversarial attack is more realistic, which is correspond to our speculation. 

\section{Discussion}
\subsection{Input from trading experts}
For now, our study collects questionnaires from the public, it will be of great advantages if we collect the response from professionals in financial industries. Owing to the high similarity between candlestick charts, the professional traders are able to analyze the candlestick data more efficiently than normal people. Although it might be hard to collect enough samples from this highly specialized group, the opinion provided by expert could help improving our models' capability of data augmentation.  

\subsection{Benefits for machine learning in Finance}
Since our model can successfully generate realistic data, it activates the interest that if it can really benefit machine learning models to be more accurate and stable in financial applications. Future research directions may include designing experiments comparing the effectiveness of training machine learning models with original and merged data, where merged data combines the original real data and the generated data. If the accuracy and stability become better, we can state that our model not only generate the data very well but also help model training. Hopefully this will in turn expand the application scenarios of machine learning techniques in the financial field. 

\section{Conclusion}
In summary, this paper propose a Modified Local Search Attack Sampling model that demonstrates superior performance in generating candlestick data. Besides the current tools, we provide a feasible time-series data augmentation method in the field of financial trading. According to the result, the model not only precisely capture and reconstruct the specific pattern, but also produce realistic data that could confuse human. This work is of great benefit to the application of machine learning in financial sectors. We provide an open-source implementation and training data for the paper in the following URL: \url{https://github.com/pecu/FinancialVision}.

\section*{Acknowledgment}
Jun-Hao Chen and Yun-Cheng Tsai are supported in part by the Ministry of Science and Technology of Taiwan under grant 108-2218-E-002-050-. Samuel Yen-Chi Chen is supported in part by the U.S. Department of Energy, Office of Science, Office of High Energy Physics program under Award Number DE-SC-0012704 and the Brookhaven National Laboratory LDRD \#20-024. Yun-Cheng Tsai and Samuel Yen-Chi Chen conceived of the presented idea. Chia-Ying Tsao and Jun-Hao Chen developed the theory and performed the computations. All authors verified the analytical methods and discussed the results and contributed to the final manuscript. Thanks to Prof. Jane Yung-Jen Hsu for constructive discussion and great support.

\bibliographystyle{unsrt}
\bibliography{references}

\end{document}